\documentclass[sigconf,nonacm]{acmart}
\usepackage{amsmath}

\AtBeginDocument{%
  \providecommand\BibTeX{{%
    \normalfont B\kern-0.5em{\scshape i\kern-0.25em b}\kern-0.8em\TeX}}}

\setcopyright{cc} %modification --> ajout cc
\setcctype[4.0]{by}
\copyrightyear{2025}
\acmConference[XAI4Science2026 EDBT/ICDT]{International Conference on Extending Database Technology}{March 03,
  2026}{Tampere, Finland}
\begin{document}

\title{Challenges of Explainability in Continual Learning for Time Series Forecasting}


% \affiliation{%

\author{Quentin Besnard}
\email{quentin.besnard@univ-tours.fr}
\affiliation{%
  \institution{LIFAT - Université de Tours}
  \city{Tours}
  \country{France}
}

\author{Emmanuel Doumard}
\email{emmanuel.doumard@univ-tours.fr}
\affiliation{%
  \institution{LIFAT - Université de Tours}
  \city{Tours}
  \country{France}
}

\author{Nicolas Labroche}
\email{nicolas.labroche@univ-tours.fr}
\affiliation{%
  \institution{LIFAT - Université de Tours}
  \city{Tours}
  \country{France}
}

\author{Nicolas Ragot}
\email{nicolas.ragot@univ-tours.fr}
\orcid{0000-0003-2321-942X}
\affiliation{%
  \institution{LIFAT - Université de Tours}
  \city{Tours}
  \country{France}
}

\author{Nicolas Ringuet}
\email{nicolas.ringuet@univ-tours.fr}
\affiliation{%
  \institution{LIFAT - Université de Tours}
  \city{Tours}
  \country{France}
}

\renewcommand{\shortauthors}{Besnard, et al.}

\begin{abstract}
Deep learning models have shown strong potential for time series forecasting, yet their deployment in real-world environmental monitoring remains challenging due to non-stationary dynamics and limited explainability. In this work, we investigate explainability as a central tool for understanding continual learning in adaptive time series forecasting, with Experience Replay strategies. We study neural forecasting architectures such as PatchMixer, PatchTST and DLinear, augmented with attention-based sampling mechanisms to support model adaptation over time. Explainability is leveraged through attention rollout and gradient-based attribution methods (Grad-CAM) to analyze both predictive behavior and sampling strategies within a continual learning framework. Experiments conducted on real-world piezometric time series exhibiting heterogeneous patterns and regime shifts show that analyzing model and sampling behaviors provides valuable insights into the dynamics of the continual learning framework. Beyond predictive performance, our results highlight the challenges and opportunities of using explainability to understand continual learning behaviors, revealing how attribution patterns evolve over time and how they can inform data selection and adaptation strategies in non-stationary forecasting scenarios.
\end{abstract}

%%
%% The code below is generated by the tool at: http://dl.acm.org/ccs.cfm
\begin{CCSXML}
<ccs2012>
<concept>
<concept_id>10010147.10010257.10010293.10010294</concept_id>
<concept_desc>Computing methodologies~Neural networks</concept_desc>
<concept_significance>300</concept_significance>
</concept>
</ccs2012>
\end{CCSXML}

\ccsdesc[300]{Computing methodologies~Neural networks}

\keywords{continual learning, time series forecasting, non-stationary environment, explainability, attention, GWL prediction}

\maketitle

\section{Introduction} %modification --> reprise de l'introduction pour mieux introduire le papiersous la vision de "Challenges of"
Recent advances in deep learning have led to significant progress in time series forecasting across a wide range of applications, from energy systems to climate and environmental monitoring. Architectures based on convolutional and transformer mechanisms have demonstrated strong predictive capabilities on both short- and long-term horizons. However, despite these advances, deploying such models in real-world environmental settings remains challenging due to the inherently non-stationary nature of the data and the limited interpretability of learned representations.

In hydrogeological contexts such as piezometric level forecasting, temporal dynamics are influenced by complex and evolving anthropic factors including seasonal variability extreme climatic events, and long-term effects. These factors induce distribution shifts and regime changes that can severely degrade the performance of forecasting models. Continual learning offers a promising paradigm to address this issue by enabling models to incrementally adapt to new data while preserving previously acquired knowledge. Yet, in practice, continual learning for time series forecasting remains relatively underexplored, particularly with respect to understanding how adaptation mechanisms behave over time and under non-stationarity.

At the same time, explainability has emerged as a critical requirement for deploying deep learning models in environmental decision-making systems. Stakeholders require interpretable insights to assess model reliability, validate predictions against domain knowledge, and understand the influence of historical observations on future forecasts. Beyond transparency and trust, explainability provides analytical tools to inspect model behavior, assess the influence of historical observations, and relate predictions to domain knowledge. However, when combined with continual learning, explainability raises new challenges: attribution signals may evolve over time, attention mechanisms may shift across regimes, and the interpretation of memory usage becomes increasingly complex as models adapt.

In this work, we investigate explainability as a tool for analyzing and understanding continual learning in adaptive time series forecasting. Rather than proposing a definitive solution, we adopt an exploratory perspective based on experiments. We focus on neural forecasting architectures commonly adopted in recent literature, including PatchMixer, PatchTST, and DLinear, and extend them with continual adaptation through Experience Replay and Knowledge Distillation. Several replay sampling strategies are considered, including random and loss-based sampling, as well as an attention-based strategy that leverages self-attention mechanisms to identify informative temporal segments

To analyze both predictive behavior and memory management strategies, we leverage attention rollout within the sampling mechanism and gradient-based attribution methods (Grad-CAM) applied to the forecasting model. These explainability signals are used not only to interpret model predictions, but also to study why specific sequences are selected for replay and how they contribute to knowledge retention and adaptation across learning phases.

We evaluate our approach on real-world piezometric time series exhibiting heterogeneous temporal patterns and pronounced regime shifts. Experimental results show that attention-guided Experience Replay strategies can help reveal several open challenges related to the interpretation of attribution signals, the stability of explanations over time, and their use for guiding adaptation. Beyond performance improvements, our analysis provides novel insights into the internal dynamics of continual learning for time series forecasting and provides a foundation for discussing future research directions and open problems at the intersection of explainability, continual learning, and non-stationary time series forecasting.

\section{Motivation and Related Work}
Deep learning–based approaches have become increasingly prominent in time series forecasting, achieving state-of-the-art performance across a wide range of applications, including energy systems, finance, and environmental monitoring. Recent architectures such as PatchTST \cite{Nie2022} and PatchMixer \cite{Gong2023} leverage patch-based representations to efficiently model long-range temporal dependencies, either through self-attention mechanisms or convolutional mixing strategies. For comparison, we also include DLinear \cite{Zeng2022}, a simpler baseline model based on a linear decomposition layer. While these models demonstrate strong predictive capabilities under stationary settings, their deployment in real-world environmental systems remains challenging due to persistent non-stationarity, regime shifts, and evolving data distributions \cite{Gunasekara2023}.

In hydrogeological applications, such as piezometric level forecasting \cite{Chen2023}, groundwater dynamics are influenced by long-term climatic trends, episodic droughts, and changing recharge patterns. These factors induce gradual drifts as well as abrupt changes in the statistical properties of the observed time series, rendering static training paradigms inadequate. Continual learning has therefore emerged as a promising framework to address these challenges by enabling models to incrementally adapt to new data while preserving previously acquired knowledge \cite{Gupta2022}. Among continual learning strategies, Experience Replay are widely adopted to mitigate catastrophic forgetting, particularly in sequential and streaming settings.

A critical yet often overlooked component of continual learning systems is the data sampling strategy used to populate and update the replay buffer. Existing approaches range from random sampling to heuristic criteria based on prediction loss or uncertainty \cite{Schillaci2021, Smith2024}. While loss-based strategies can improve adaptation to novel regimes, they provide limited explainability and may bias the learning process toward extreme or noisy samples. Recent works have begun to explore attention-based mechanisms \cite{Sokar2021,Cai2023,Xue2022} for data selection, leveraging internal model representations to identify informative samples in a more structured manner. However, the explanation of such sampling decisions, especially in the context of time series forecasting, remains insufficiently studied.

In parallel, the field of explainable artificial intelligence (XAI) \cite{Rojat2021} has developed a rich set of tools to interpret neural networks, including gradient-based attribution methods such as Grad-CAM and attention visualization techniques. While these methods are commonly used to analyze predictive models, their integration into continual learning pipelines, particularly as a means to interpret and guide data selection strategies, has received limited attention \cite{proietti2025xai}. Bridging this gap is especially important in environmental applications, where trust, transparency, and explanation are essential for scientific understanding and decision support. Some example with XAI has already shown promising results to guide model training through interpretable regularization to reduce concept drift in class incremental continual learning \cite{Rymarczyk_2023_ICCV}.

Motivated by these limitations, this work investigates the role of explainability in understanding and guiding continual learning for non-stationary time series forecasting. By jointly analyzing attention-based sampling mechanisms and model-level attribution signals, we aim to provide a data-centric perspective on continual adaptation, highlighting how explainability can help interpret the selection of replay samples and elucidate the internal dynamics of adaptive forecasting models.

\section{Piezometric Datasets}
In the current context of climate change, characterized by shifts in precipitation regimes, increasing evapotranspiration, and a higher frequency of drought events, long-term analyses of piezometric time series provide valuable insights into the evolving behavior of groundwater systems. Investigating multi-decadal trends in groundwater levels enables the detection of distributional changes and structural modifications in aquifers subjected to increasingly variable climatic forcing. In this study, we focus specifically on two drastically different piezometers selected from a subset of 17 groundwater-level datasets collected at piezometric monitoring stations across France\footnote{Datasets are available at \url{https://hubeau.eaufrance.fr/}, with access to over 1,500+ piezometers.}. 

\begin{figure}[!ht]
    \centering
    \includegraphics[width=\linewidth]{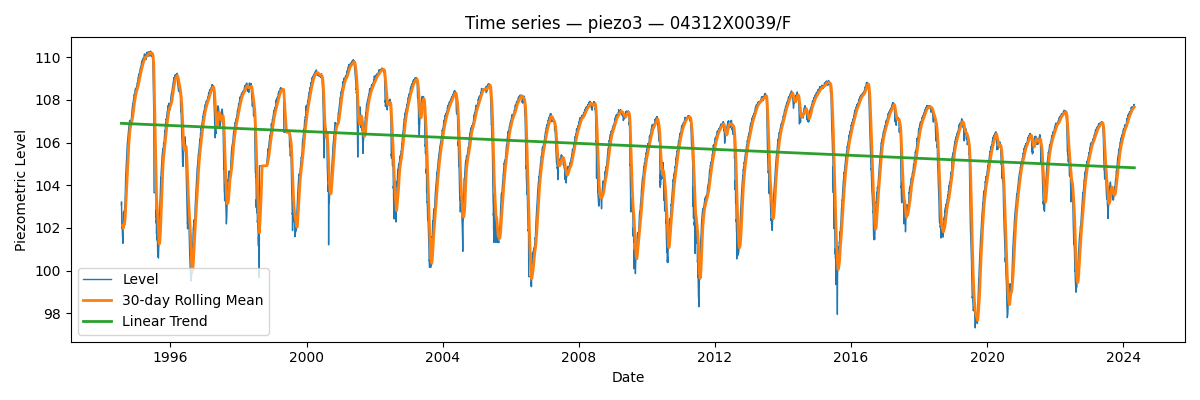}
\caption{Groundwater level time series and associated linear trend for Piezometer 3 (Beauce region).}%modification --> titre figure
\label{fig:piezo_ts1}
\end{figure}

\begin{figure}[!ht]
    \centering
    \includegraphics[width=\linewidth]{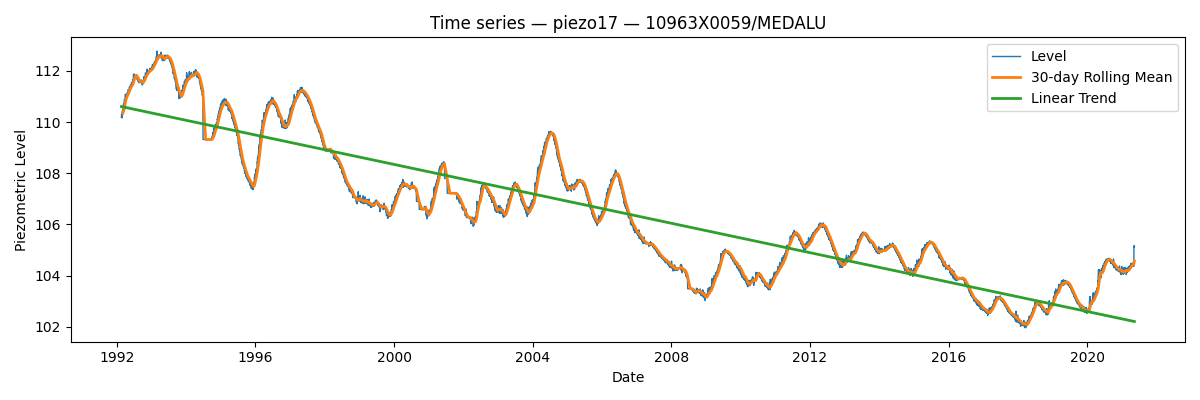}
\caption{Groundwater level time series and associated linear trend for Piezometer 17 (Pyrénée-Orientales region).}%modification --> titre figure
\label{fig:piezo_ts2}
\end{figure}

Figures \ref{fig:piezo_ts1} and \ref{fig:piezo_ts2} display the temporal evolution of groundwater levels for two representative piezometers, along with their associated linear trends. These time series illustrate a clear modification of groundwater dynamics over time, with noticeable contrasts between the early and recent periods. The seasonal extrema, associated with recharge events and low-water conditions, exhibit increasing irregularity, departing from a strictly annual cycle. This loss of seasonal consistency reflects enhanced variability in recharge processes and a gradual disruption of historical hydrogeological patterns.

\begin{figure}[!ht]
    \centering
    \includegraphics[width=\linewidth]{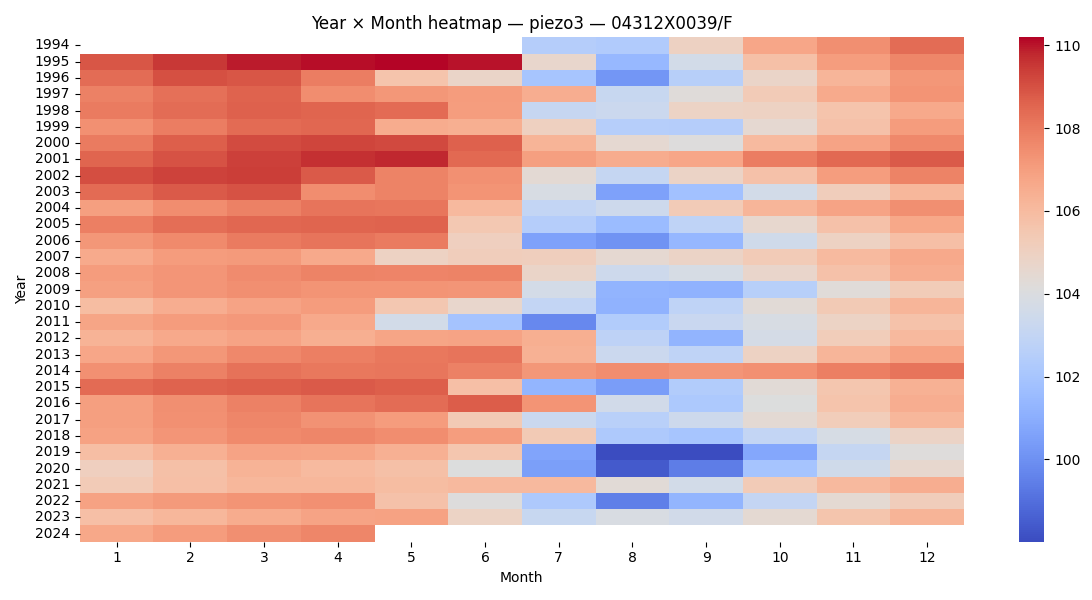}
\caption{Heatmap of average monthly groundwater level over multiple decades (x-axis: months; y-axis: years). Example shown for Piezometer 3 (Beauce region).}%modification --> titre figure
\label{fig:piezo_heatmap1}
\end{figure}

\begin{figure}[!ht]
    \centering
    \includegraphics[width=\linewidth]{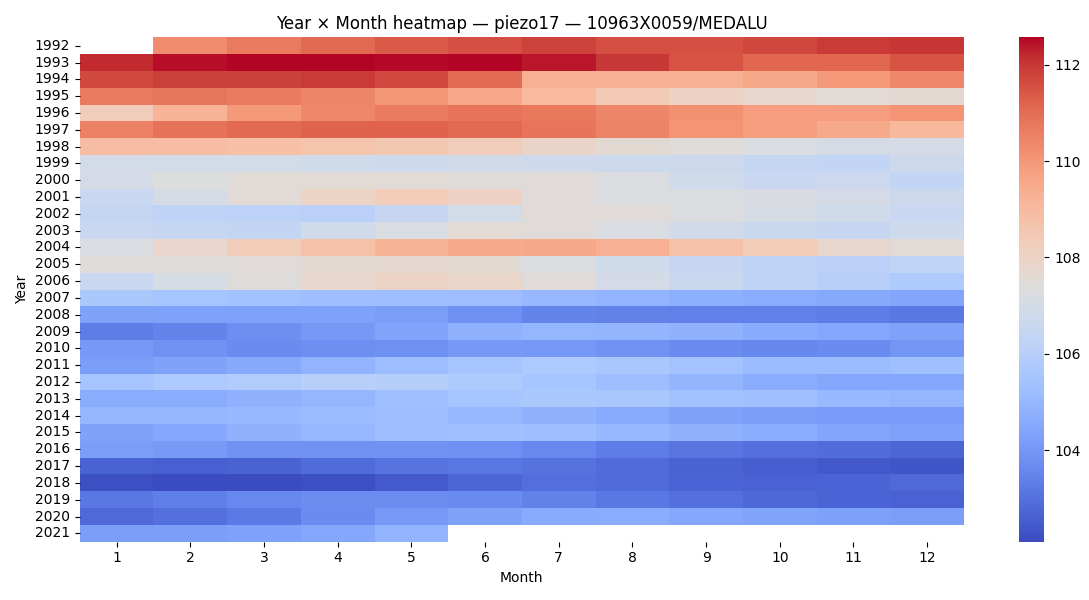}
\caption{Heatmap of average monthly groundwater level over multiple decades (x-axis: months; y-axis: years). Example shown for Piezometer 17 (Pyrénée-Orientales region).}%modification --> titre figure
\label{fig:piezo_heatmap2}
\end{figure}

The heatmaps of average monthly groundwater levels (Figures \ref{fig:piezo_heatmap1} and \ref{fig:piezo_heatmap2}), computed over more than twenty years, further emphasize the presence of long-term drift. While the 1990s and early 2000s are characterized by relatively elevated water tables and well-defined winter recharge phases, subsequent years reveal a progressive lowering of average levels. From the 2010s onward, summer low-water periods become more pronounced, and winter recovery weakens, pointing to a sustained downward trend that cannot be explained solely by seasonal or interannual variability.

\section{Implementation} %modification --> reprise pour donner un peu plus de détail sur le setup expérience
All experiments address long-term multivariate time series forecasting under a continual learning setting. The model is first trained on 50\% of the data, with an additional 10\% used as a validation subset, and is subsequently adapted as new data arrive sequentially through a replay-based memory mechanism. For long-term forecasting, the input sequence length is set to \texttt{seq\_len = 180} and the prediction horizon to \texttt{pred\_len = 360}, corresponding to approximately one year of daily piezometric observations. For the piezometric datasets, incomplete records are removed and occasional missing values are filled using linear interpolation. No manual correction of outliers is applied in order to preserve realistic deployment conditions.

We adopt a model-agnostic continual learning framework that combines knowledge distillation with replay-based memory management. Different replay sampling strategies are considered, including random, loss-based, and attention-based sampling. Model updates are triggered at predefined adaptation frequencies. At each update, a student model is trained via knowledge distillation from the current teacher using only the samples stored in the replay buffer. Once training is completed, the student replaces the teacher and becomes the new predictor for subsequent adaptation phases. Experiments are conducted using PatchTST \cite{Nie2022}, PatchMixer \cite{Gong2023}, and DLinear \cite{Zeng2022}, which rely on distinct architectural paradigms.

All predictor models are trained using the same optimization procedure and hyperparameters across experiments: \texttt{hidden\_size = 256}, \texttt{lr = 1e-3}, \texttt{epochs = 100}, \texttt{batch\_size = 16}, with early stopping controlled by a validation subset (\texttt{patience = 3}). Continual adaptation is performed with a fixed update frequency of \texttt{freq = 300} and a replay buffer of size \texttt{buffer\_size = 1500}. For the attention-based strategy, which includes an additional sampling module, an inner-loop parameter of \texttt{inner\_loop = 3} is used. All experiments are run using CUDA on a dedicated server equipped with an Intel(R) Xeon(R) Gold 5220R CPU (2.20\,GHz) and an NVIDIA RTX A6000 GPU.

For the analysis of adaptation and memory dynamics, each data sample is assigned a unique identifier to track its replay frequency over time. Attention rollout is used within the attention-based sampling strategy to record sample importance scores, while model-level attribution is obtained using Grad-CAM applied to the final linear layer of the forecasting models. These explainability signals are collected throughout training and adaptation to analyze the relationship between replay selection, attribution stability, and continual learning behavior.

\section{Analysis}
While the proposed continual learning framework consistently yields performance improvements, understanding how data are selected for replay and which temporal regions are used by the forecasting models is essential for interpreting these gains and assessing their robustness. The analysis will focus specifically on Piezometer 3 to provide a more detailed case study, and the following forecasting figure shows only the testing data subset.

\begin{figure}[!ht]
    \centering
    \includegraphics[width=\linewidth]{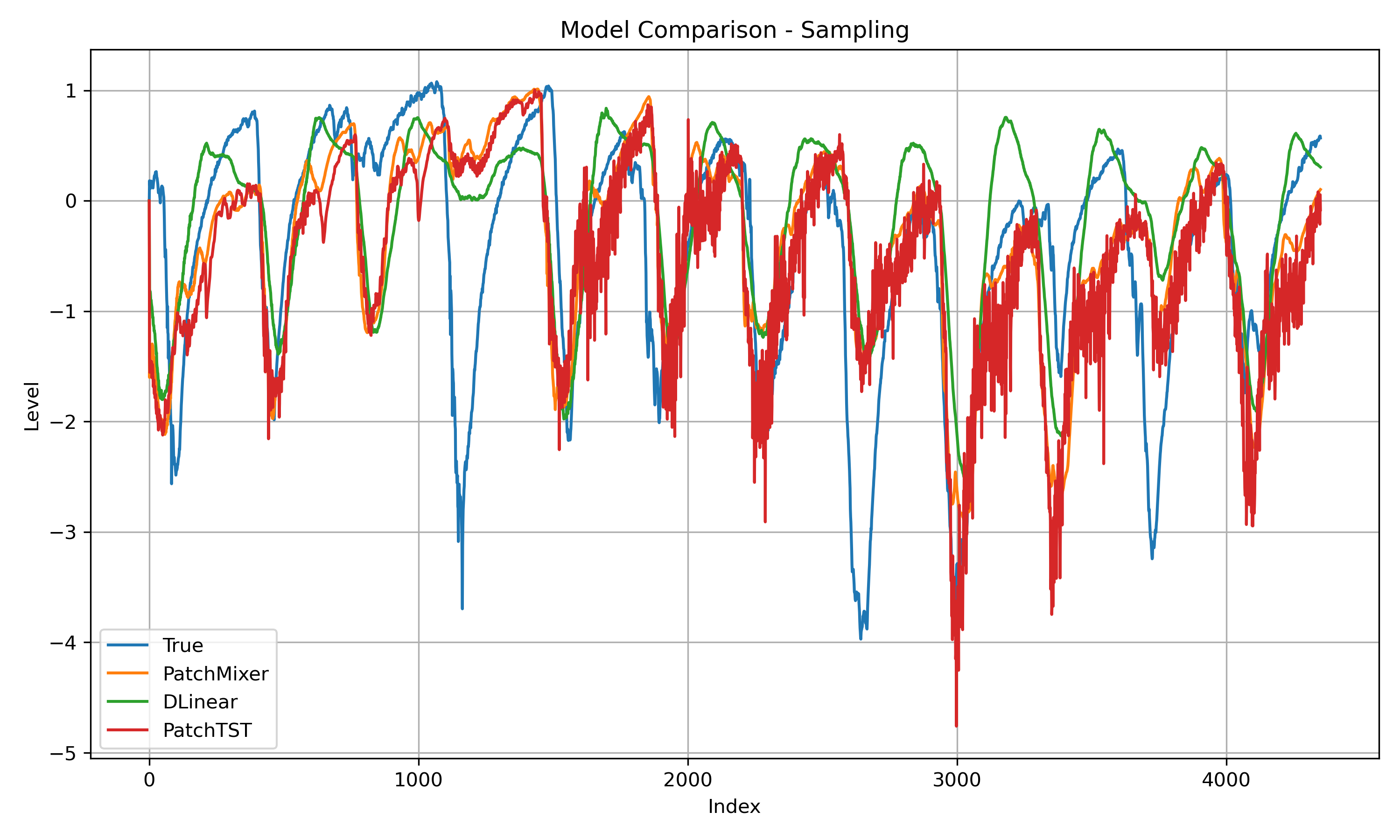}
\caption{Continual learning framework with different forecasting models comparison applied on piezometer 3 with Attention Experience Replay strategy.} %modification --> + "with Attention Experience Replay strategy"
\label{fig:comparaison_standard_kd2}
\end{figure}

Figure \ref{fig:comparaison_standard_kd2} presents the forecasting performance on the test set of Piezometer 3, obtained using a continual learning setup integrating both distillation and attention mechanisms. The results confirm that attention-guided replay contributes to improved predictive accuracy under non-stationary conditions. However, performance metrics alone do not provide insights into the internal decision processes driving these improvements.

\begin{figure}[!ht]
    \centering
    \includegraphics[width=\linewidth]{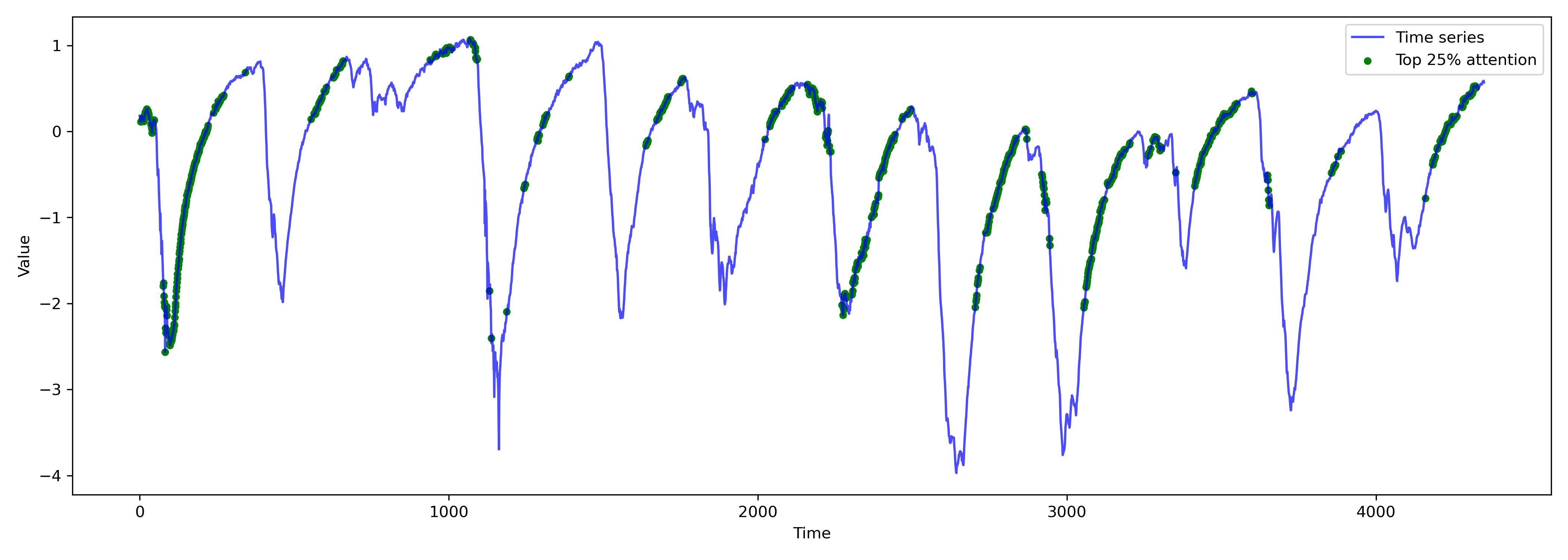}
\caption{PatchMixer Grad-CAM focus points.}
\label{fig:focus_points1_3}
\end{figure}

\begin{figure}[!ht]
    \centering
    \includegraphics[width=\linewidth]{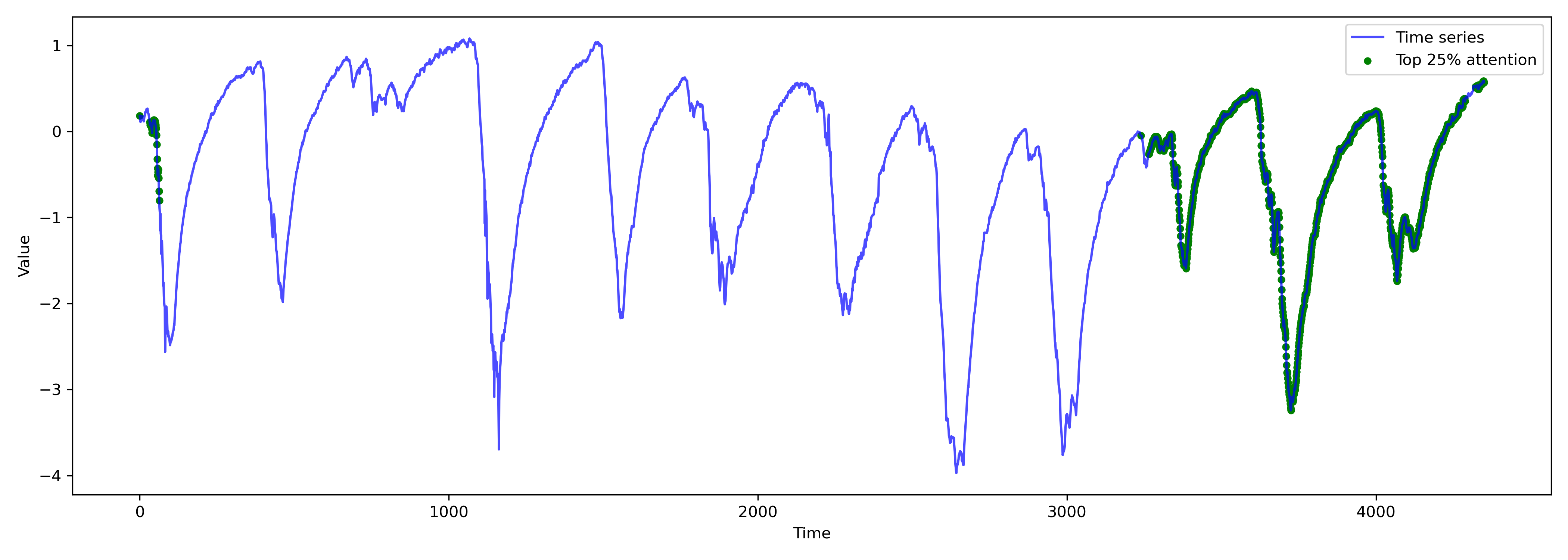}
\caption{PatchTST Grad-CAM focus points.}
\label{fig:focus_points3_3}
\end{figure}
To address this limitation, we analyze the explainability signals produced by both the forecasting models and the attention-based sampling mechanism. Figures \ref{fig:focus_points1_3} and \ref{fig:focus_points3_3} illustrate the Grad-CAM attribution maps for the PatchMixer and PatchTST forecasting models applied to Piezometer 3. The green dots (and red dots for the Attention Rollout) indicate the top 25\% highest-weighted data points, aggregated over time.

The Grad-CAM analyses reveal substantial architectural differences in how the two models exploit temporal information. PatchMixer, whose core relies on convolutional operations, displays a relatively uniform attribution across time, suggesting a more distributed use of historical observations. Conversely, PatchTST, which is based on a transformer architecture, exhibits a pronounced concentration of attributions toward the most recent time steps. This behavior reflects the combined effects of self-attention mechanisms and positional encoding, which tend to amplify the contribution of recent observations and introduce a temporal bias toward the end of the input sequence.

\begin{figure}[!ht]
    \centering
    \includegraphics[width=\linewidth]{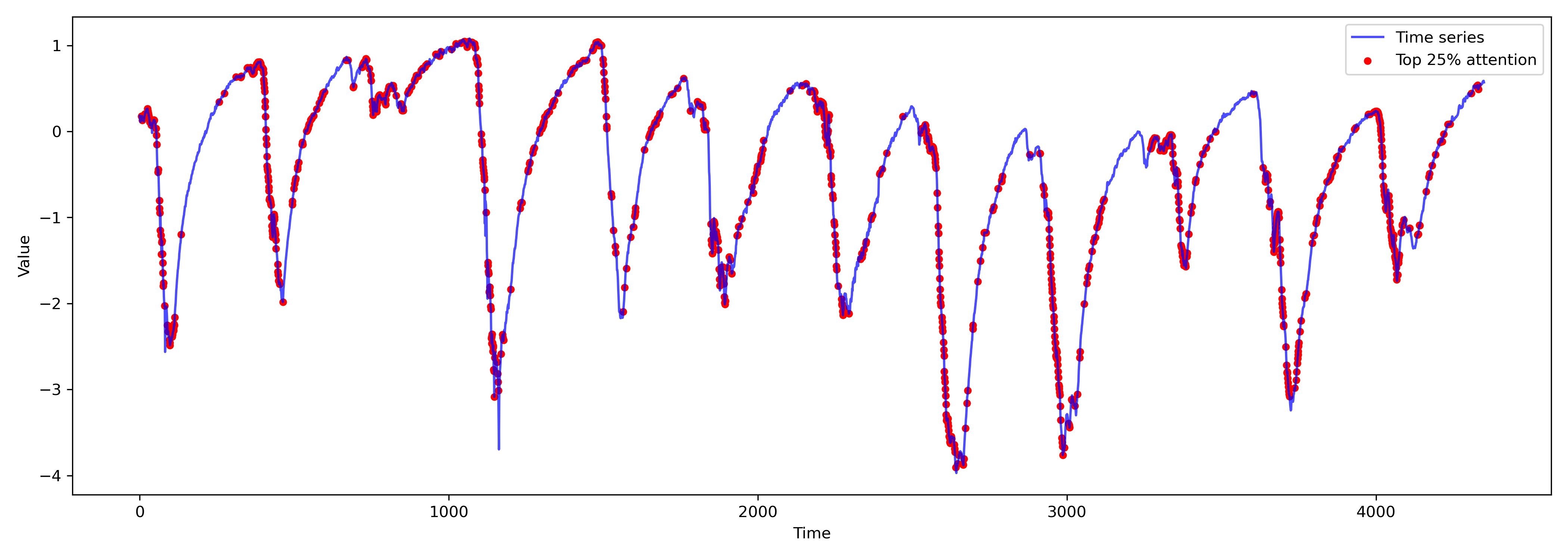}
\caption{PatchMixer Attention Rollout focus points.}
\label{fig:focus_points2_3}
\end{figure}

\begin{figure}[!ht]
    \centering
    \includegraphics[width=\linewidth]{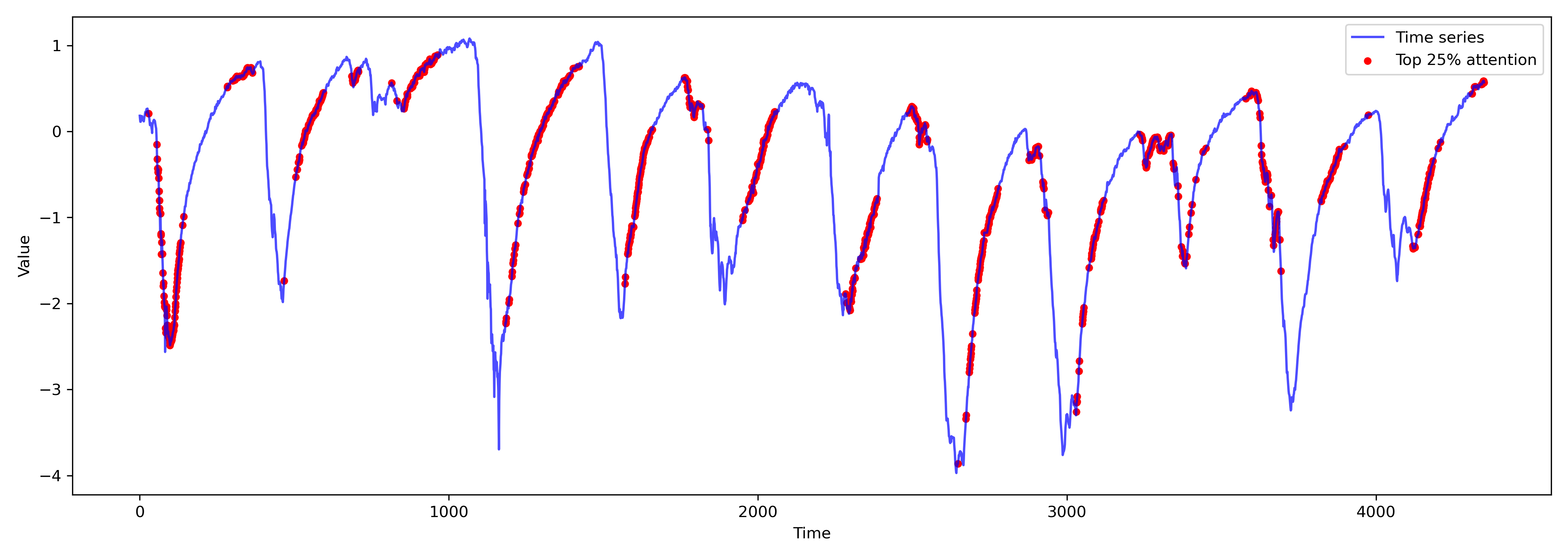}
\caption{PatchTST Attention Rollout focus points.}
\label{fig:focus_points4_3}
\end{figure}
In contrast, Figures~\ref{fig:focus_points2_3} and~\ref{fig:focus_points4_3} present the corresponding Attention Rollout visualizations for the PatchMixer and PatchTST models.

The Attention Rollout results exhibit consistent selection patterns across PatchMixer and PatchTST. This consistency suggests that the attention-based sampling mechanism learns stable, model-agnostic criteria for identifying informative temporal segments, despite differences in the underlying forecasting architectures. Several temporal regions are similarly emphasized across both models. By contrast, the primary distinction between the two configurations lies in the forecasting model used for gradient propagation during learning, rather than in the sampling strategy itself.

\begin{figure}[!ht]
    \centering
    \includegraphics[width=\linewidth]{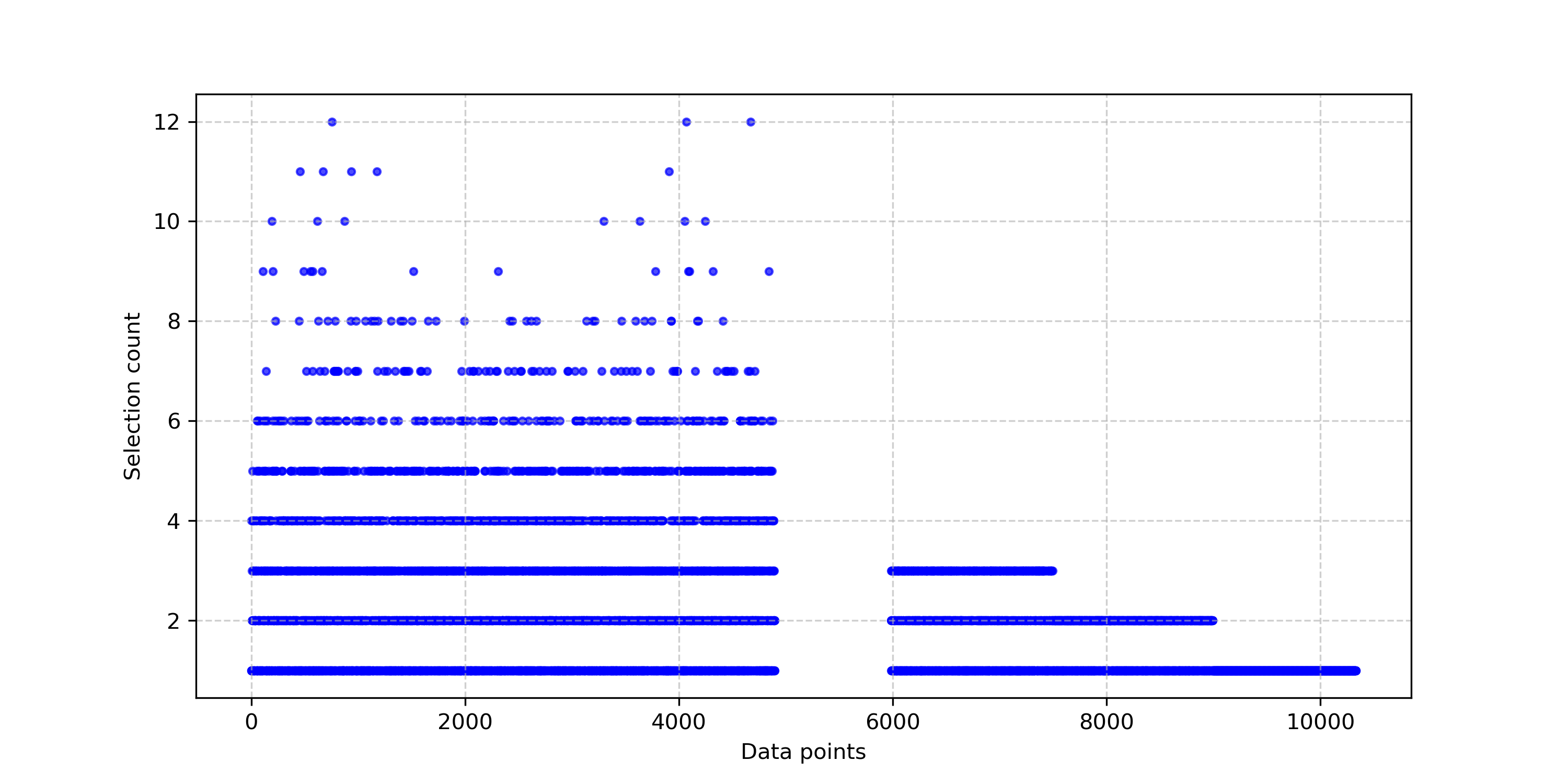}
\caption{Random sampling strategy.}
\label{fig:four_images1}
\end{figure}

\begin{figure}[!ht]
    \centering
    \includegraphics[width=\linewidth]{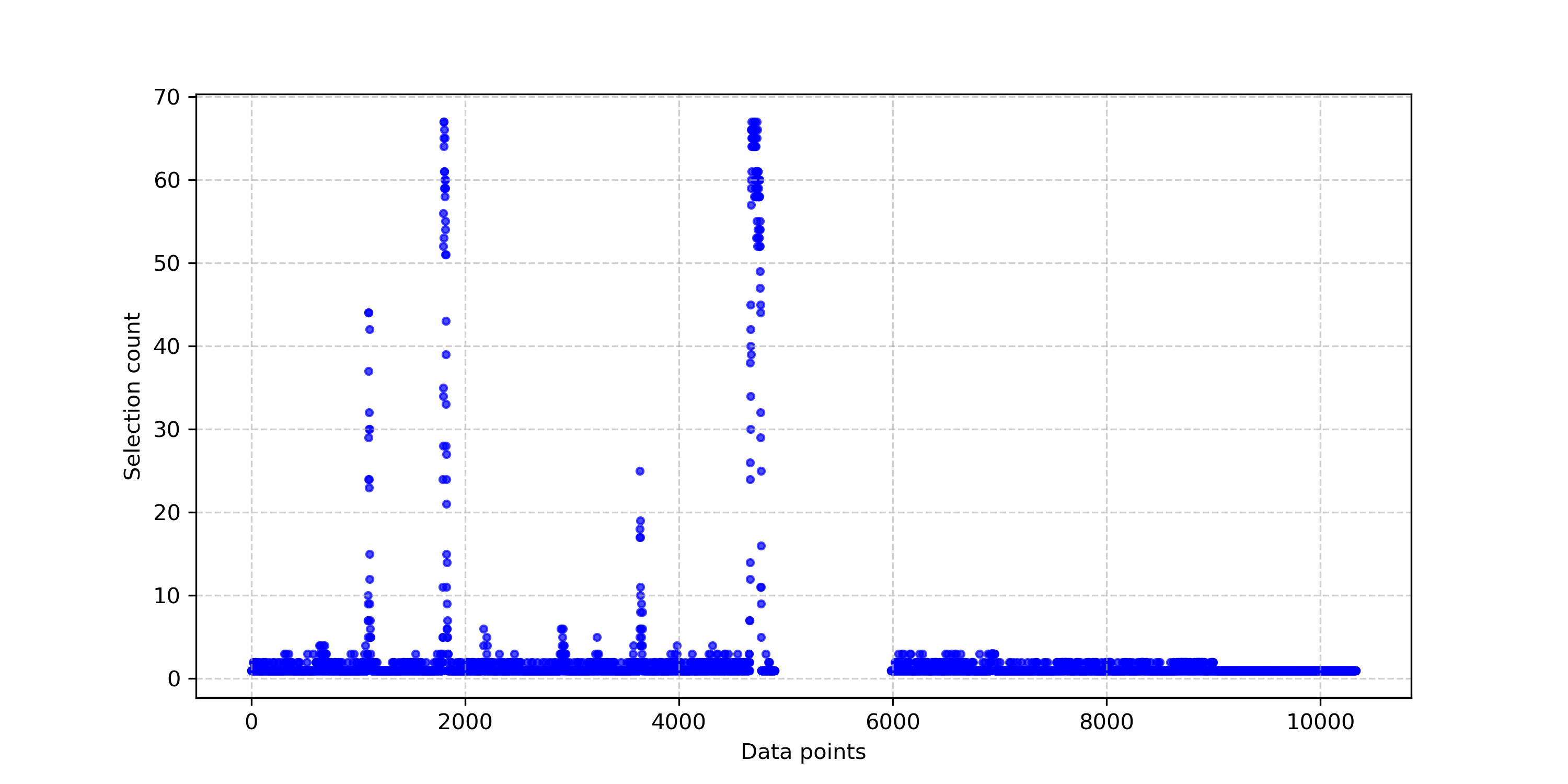}
\caption{Attention sampling strategy.}
\label{fig:four_images2}
\end{figure}

\begin{figure}[!ht]
    \centering
    \includegraphics[width=\linewidth]{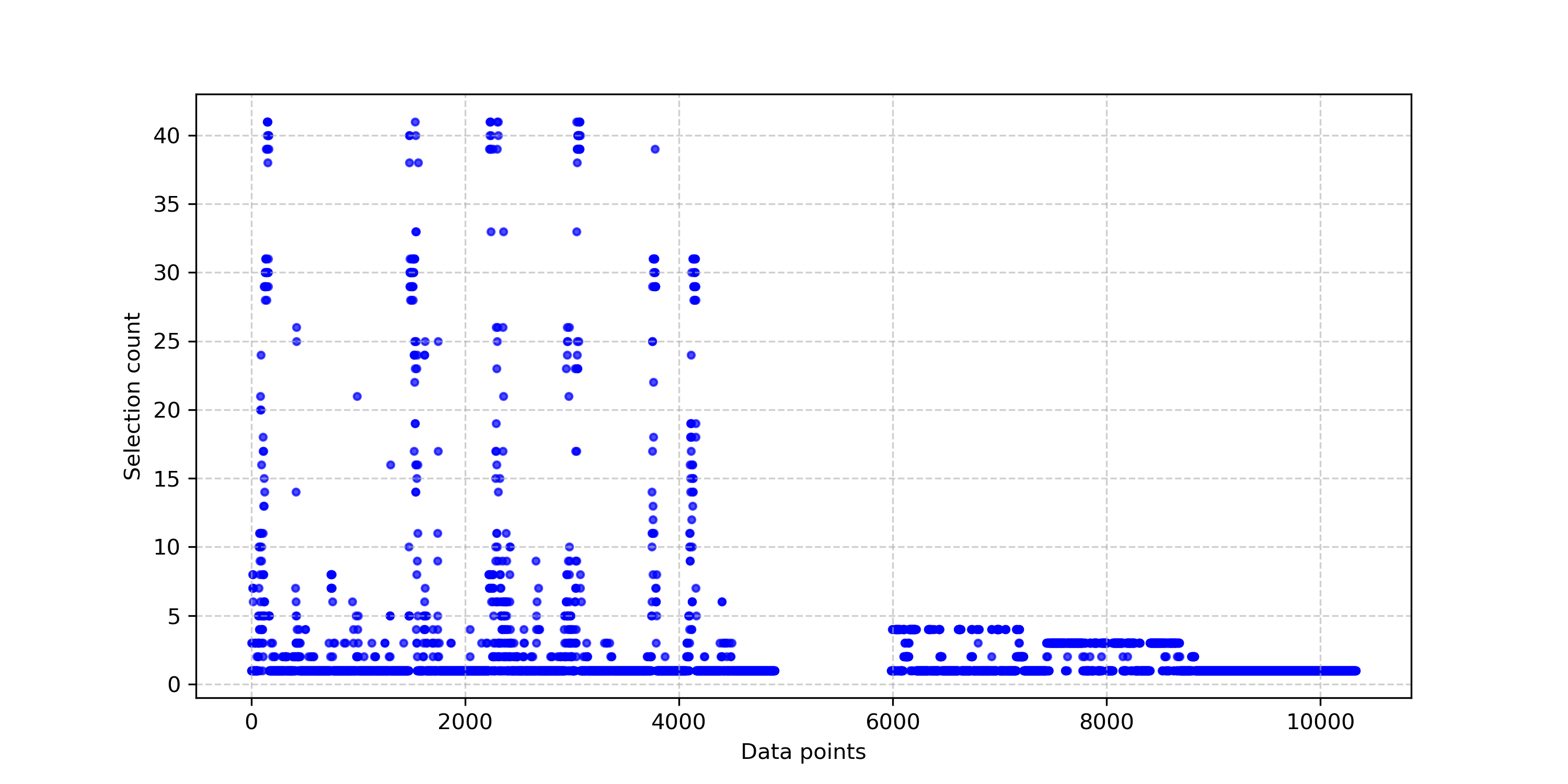}
\caption{Maximum loss sampling strategy.}
\label{fig:four_images3}
\end{figure}

\begin{figure}[!ht]
    \centering
    \includegraphics[width=\linewidth]{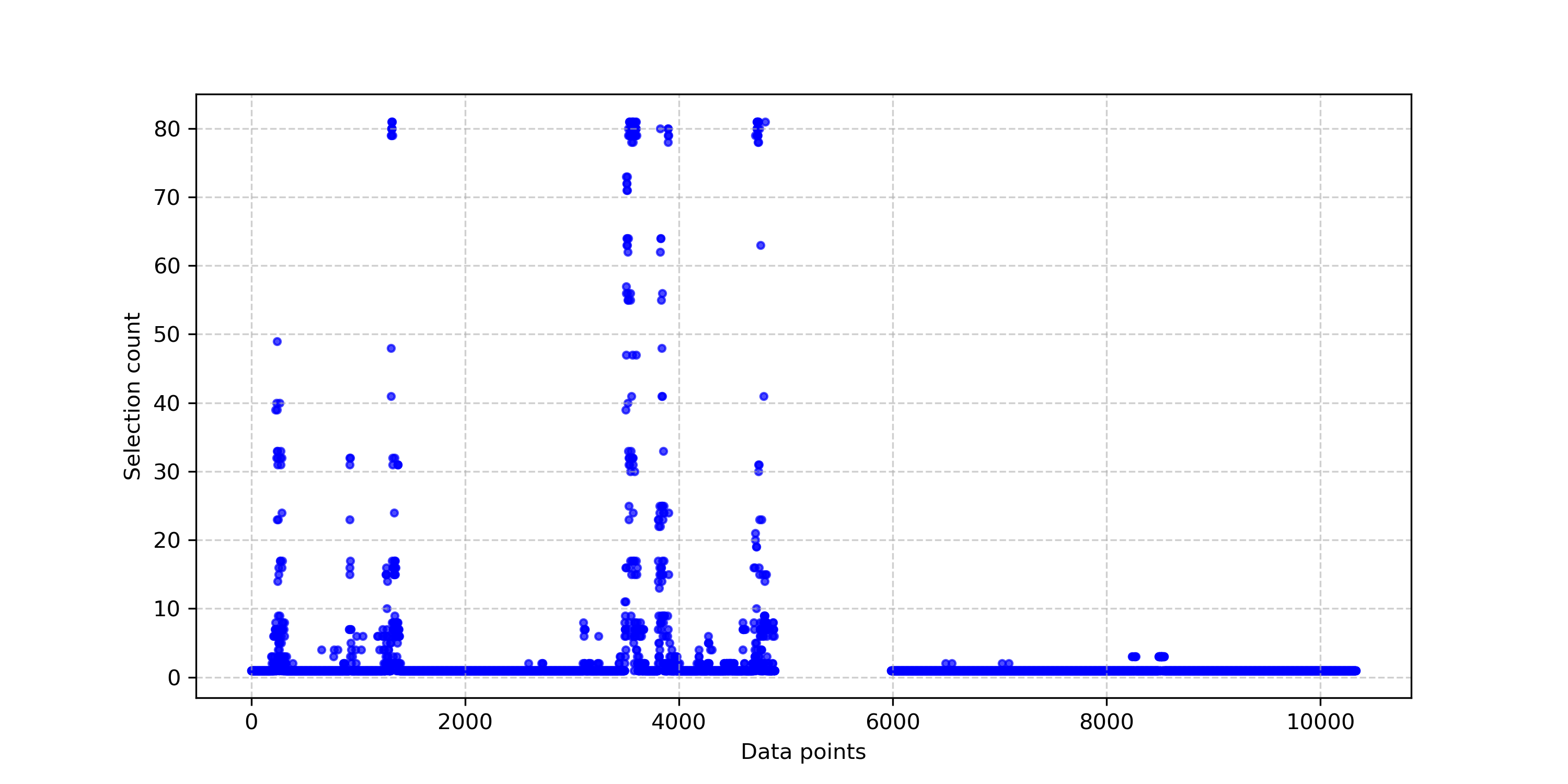}
\caption{Minimum loss sampling strategy.}
\label{fig:four_images4}
\end{figure}

Figures \ref{fig:four_images1}, \ref{fig:four_images2}, \ref{fig:four_images3} \& \ref{fig:four_images4} illustrate the temporal sampling patterns induced by the different data selection strategies evaluated in this study. For each strategy, the indices of the sampled sequences are recorded in order to analyze how the selection process evolves over time. A clear discontinuity is observed between indices 5000 and 6000, corresponding to the validation split. Since no learning occurs on this subset, samples selected before this interval belong to the training phase, whereas those selected afterward correspond to the test data, on which the predictive model is incrementally adapted.

The sampling density is substantially higher within the training region, as this phase aggregates samples collected over multiple epochs under a standard batch-learning regime. However, the most informative behavior emerges during the testing stage, where continuous sampling of new observations governs the model’s capacity for adaptation. In this context, the sampling strategy plays a critical role in determining which data points are deemed informative and therefore contribute to the continual update of the predictive model.

Random sampling (Figure \ref{fig:four_images1}) yields a uniform and temporally unstructured distribution of selected samples. While this strategy ensures unbiased coverage of the data stream, it does not provide any explicit insight into the relevance or contribution of individual samples for model adaptation. Consequently, from an explainability perspective, the selection process remains opaque, as all data points are treated as equally informative regardless of their impact on the model’s internal representations or prediction error.

In contrast, the attention-based sampling strategy (Figure \ref{fig:four_images2}) exhibits a more structured and heterogeneous selection pattern. By leveraging attention weights, this approach prioritizes samples that are deemed more influential by the model’s internal mechanisms. This behavior enhances data-level explainability, as the sampling decisions can be directly linked to learned importance scores, providing an interpretable rationale for why specific observations are retained for continual adaptation. Such a mechanism facilitates the identification of critical temporal segments and promotes adaptive learning driven by semantically meaningful data.

The maximum-loss sampling strategy (Figure \ref{fig:four_images3}) further emphasizes difficult or poorly predicted samples, leading to a concentrated selection of high-error instances. While this strategy can accelerate adaptation to novel or challenging regimes, it offers a more limited form of explainability, as the selection criterion is solely based on error magnitude and does not capture the broader contextual or structural relevance of the data. Moreover, excessive focus on extreme errors may increase sensitivity to noise or transient anomalies.

Finally, the minimum-loss sampling strategy (Figure \ref{fig:four_images4}) highlights a key limitation in the context of continual adaptation. By repeatedly selecting samples associated with low prediction error, this approach favors “easy” instances and results in a highly redundant sampling pattern. Such behavior restricts exploration of informative or novel data and may ultimately hinder adaptation by reinforcing existing biases. From an explainability standpoint, this strategy provides little insight into the drivers of model change, as it systematically overlooks samples that could challenge or enrich the learned representations.

Overall, these results demonstrate that sampling strategies grounded in model-derived importance measures, such as attention-based selection, offer a more interpretable and effective mechanism for guiding continual adaptation. Thanks to the discriminative power provided by attention, new elements are continually explored while preserving the relevance of past observations, as reflected in the sampling process. By explicitly linking data selection to internal model dynamics, such strategies enhance both the transparency of the learning process and the robustness of the downstream predictive model.

\section{Conclusion} 
This study investigates the interplay between continual learning, data sampling strategies, and explainability in the context of non-stationary time series forecasting for groundwater monitoring, addressing challenges highlighted in the introduction. By integrating attention-based sampling, adaptive replay mechanisms can effectively improve long-term forecasting performance under evolving hydrogeological conditions, demonstrating how continual learning can mitigate the effects of distribution shifts and regime changes.

Beyond predictive accuracy, our analysis emphasizes the central role of explainability for understanding how continual learning systems adapt over time. Attention Rollout and buffer sampling analyses provide insight into why specific sequences are selected for replay, revealing stable and model-agnostic sampling patterns driven by learned importance scores. In parallel, Grad-CAM highlights how different forecasting architectures exploit temporal information, exposing distinct inductive biases between convolution-based and transformer-based models, as anticipated in our discussion of architectural differences in the introduction.

The comparative analysis of buffer sampling strategies further demonstrates the importance of approaches that strike a favorable balance between newly observed and previously learned data points. Purely loss-based or random sampling strategies, by contrast, either obscure the rationale behind data selection or risk reinforcing biased or redundant learning behaviors. These findings underscore the value of explainability as both a diagnostic and analytical tool for data-centric continual learning, directly addressing the interpretability challenges highlighted at the outset.

Overall, this work contributes to understanding continual adaptation in time series forecasting and the challenges of implementing explainability in real-world applications, where explainability is crucial for interpreting both predictions and internal learning dynamics. These insights are particularly relevant for environmental monitoring, where non-stationarity is inherent and transparent decision-making is essential. Future work will focus on extending these principles to additional features and environmental variables, and on investigating how explainability can guide the design of sampling strategies that are further optimized for robustness and long-term generalization, thereby strengthening the link between model transparency and practical deployment in non-stationary settings.

\bibliographystyle{ACM-Reference-Format}
\bibliography{references.bib}

\end{document}